\documentclass{svproc}
\usepackage[utf8]{inputenc}

\usepackage{url}

\usepackage{amsmath,amsfonts}
\usepackage{algorithmic}
\usepackage{algorithm}
\usepackage{array}
\usepackage[caption=false,font=normalsize,labelfont=sf,textfont=sf]{subfig}
\usepackage{textcomp}
\usepackage{stfloats}
\usepackage{verbatim}
\usepackage{graphicx}
\usepackage{xcolor}
\usepackage{hyperref}

\usepackage[backend=biber, style=numeric]{biblatex}

\begin{document}
\mainmatter              
\title{Awareness in robotics: An early perspective from the viewpoint of the EIC Pathfinder Challenge ``Awareness Inside''}
\titlerunning{Awareness in robotics}  
%
\author{
%
%
Cosimo Della Santina\inst{1,2} \and Carlos Hernandez Corbato\inst{1} \and Burak Sisman\inst{1} \and
%
Luis A. Leiva\inst{3} \and Ioannis Arapakis\inst{4} \and Michalis Vakalellis\inst{5} \and Jean Vanderdonckt\inst{6} \and
%
Luis Fernando D’Haro \inst{7} \and Guido Manzi \inst{8} \and Cristina Becchio \inst{9} \and Aïda Elamrani \inst{10} \and
%
Mohsen Alirezaei\inst{11} \and Ginevra Castellano\inst{12} \and Dimos V. Dimarogonas\inst{13} \and Arabinda Ghosh\inst{14} \and Sofie Haesaert\inst{15} \and Sadegh Soudjani\inst{14} \and Sybert Stroeve\inst{16} \and
%
%
Paul Verschure\inst{17} \and 
%
Davide Bacciu\inst{18} \and Ophelia Deroy\inst{19} \and Bahador Bahrami\inst{19} \and 
Claudio Gallicchio\inst{18} \and Sabine Hauert\inst{20} \and
%
Ricardo Sanz\inst{21} \and Pablo Lanillos\inst{22, 17} \and
%
Giovanni Iacca\inst{23} \and Stephan Sigg\inst{24} \and Manel Gasulla\inst{25} \and
%
%
Luc Steels \inst{26} \and Carles Sierra \inst{27} 
}
\authorrunning{Della Santina et al.} 
%

%
\institute{Department of Cognitive Robotics, Delft University of Technology, Delft, Netherlands, {\footnotesize c.dellasantina@tudelft.nl}
\and 
Institute of Robotics and Mechatronics, German Aerospace Center (DLR), Oberpfaffenhofen, Germany.
\and 
Department of Computer Science, University of Luxembourg, Luxembourg
\and
Telefónica I+D, Spain
\and
AEGIS IT Research GmbH, Germany
\and
Université catholique de Louvain, Belgium
\and
Speech Technology and Machine Learning Group, ETSI de Telecomunicación, Universidad Politécnica de Madrid, Madrid, Spain
\and
Alien Venture Studio, Alien Technology Transfer, Rome, Italy
\and
Department of Neurology, University Medical Center Hamburg-Eppendorf (UKE), Hamburg, Germany
\and
Institut Jean Nicod, École Normale Supérieure, Paris, France
\and
Research, Technology and Development Department, Digital Industry Software, Helmond, Netherlands
\and
Department of Information Technology, Uppsala University, Uppsala, Sweden
\and
Division of Decision and Control Systems, School of Electrical Engineering and Computer Science, KTH Royal Institute of Technology, Stockholm, Sweden
\and
Max Planck Institute for Software Systems, Kaiserslautern, Germany
\and
Department of Electrical Engineering, Control Systems Group, Eindhoven University of Technology, Eindhoven, Netherlands
%
%
%
\and
Netherlands Aerospace Centre (NLR), Amsterdam, Netherlands
\and
Donders Centre for Neuroscience, Radboud University, Nijmegen, Netherlands
\and
Computer Science Department, University of Pisa, Pisa, Italy
\and
Faculty of Philosophy, Philosophy of Science and the Study of Religion, Ludwig Maximilian University of Munich, Munich, Germany
\and
School of Engineering Mathematics and Technology, University of Bristol, Bristol, United Kingdom
\and
Autonomous Systems Laboratory, Universidad Politécnica de Madrid, Spain
\and 
Cajal International Neuroscience Center (CINC), Spanish National Research Council (CSIC), Madrid, Spain
\and
Department of Information Engineering and Computer Science, University of Trento, Trento, Italy
\and
Department of Information and Communications Engineering, Aalto University, Espoo, Finland
\and
Department of Electronic Engineering, Universitat Politècnica de Catalunya, Barcelona, Spain
\and
Studio Stelluti, Brussels, Belgium
\and
Artificial Intelligence Research Institute (IIIA), Spanish National Research Council (CSIC), Barcelona, Spain
}

\maketitle              

\begin{abstract}

Consciousness has been historically a heavily debated topic in engineering, science, and philosophy. On the contrary, awareness had less success in raising the interest of scholars in the past. However, things are changing as more and more researchers are getting interested in answering questions concerning what awareness is and how it can be artificially generated. The landscape is rapidly evolving, with multiple voices and interpretations of the concept being conceived and techniques being developed. The goal of this paper is to summarize and discuss the ones among these voices that are connected with projects funded by the EIC Pathfinder Challenge called ``Awareness Inside'', a nonrecurring call for proposals within Horizon Europe that was designed specifically for fostering research on natural and synthetic awareness. In this perspective, we dedicate special attention to challenges and promises of applying synthetic awareness in robotics, as the development of mature techniques in this new field is expected to have a special impact on generating more capable and trustworthy embodied systems.

\end{abstract}

\section*{Acknowledgements}
This work is supported by the EU EIC projects: SYMBIOTIK, grant number 637107; ASTOUND, grant number {101071191}; SymAware, grant number {101070802}; CAVAA, grant number 10039052; EMERGE, grant number 101070918; METATOOL, grant number 101070940; SUST(AI)N, grant number {101071179}; and VALAWAI, grant number {101070930}. 

\section{Introduction}

According to the Cambridge English Dictionary \cite{CambridgeAwareness2024}, awareness is the ``\textit{knowledge that something exists, or understanding of a situation or subject at the present time based on information or experience}''. It may seem that, according to this broad definition, awareness is already widespread in robotics. Think, for example, of a drone creating an internal representation of its environment using a SLAM algorithm \cite{placed2023survey}, a manipulator detecting an interaction with its environment through an observer \cite{zacharaki2020safety}, or a social robot acquiring knowledge on the logical structure of its environment \cite{hernandez2018self,bruno2019knowledge} or its body~\cite{lanillos2020robot}. If we accept this broad definition, it is evident that a common framework that enables the discussion of levels of awareness across domains is missing. 
Even more, it is intuitively clear that the concept may be loaded with more complex significance than this interpretation suggests. For example, \textit{awareness} is sometimes semantically associated with the term \textit{consciousness}, which is a much more debated concept. Researchers have proposed several definitions of the latter, including philosophical \cite{tononi2015consciousness}, psychological \cite{james2007principles}, architectural \cite{dehaene2021consciousness}, neural correlates \cite{crick2003framework}, and computer science theory \cite{blum2021theoretical}.

As Artificial Intelligence (AI) progresses in reproducing a growing number of human capabilities \cite{silver2018general,salvagno2023can,epstein2023art}, it is thus natural to ask if and how synthetic awareness can be produced and embodied into physical agents. This challenge will necessarily pass through broadening awareness to include different species beyond humans and even recognizing awareness of inanimate objects. Ultimately, questions that need answering include: What is awareness exactly? Can this concept be formalized philosophically or technically? What distinguishes an aware agent from a non-aware one? What is the role of embodiment? Does awareness improve the performance of robots? Is it even ethical to endow artificial agents with awareness? 

In this vein, in June 2021, the European Innovation Council (EIC) opened a Pathfinder Challenge call for proposals with the explicit aim of pushing the boundaries of AI \cite{EICPathfinderChallenge2024}. The call challenged the scientific community to define and achieve true awareness and understand awareness beyond the perception of surroundings or self-awareness. The call also pointed towards the role that awareness could play in both clinical settings and technology, also given the importance that trustworthy AI has in this field for the EU roadmap. The call suggested that robots or decision support systems that appear `aware' could play a role in gaining human trust.

Eight projects were selected among the applicants, covering a large portion of the European research landscape: SYMBIOTIK, ASTOUND, SymAware, CAVAA, EMERGE, METATOOL, SUST(AI)N, VALAWAI. The aim of this paper is to introduce this multi-faceted view to awareness research and to provide some preliminary discussion on how these activities will impact the robotics field in the future. In the rest of the manuscript, we will review how each project looks at the challenge. In doing that, we observe that the views on awareness vary from project to project. At the same time, we show that---although not all of them are intended to generate direct applications for robotics---they all have the potential to generate disruptive innovation in this field.

\section{SYMBIOTIK: Context-aware adaptive visualizations for critical decision making}
Information Visualization (InfoVis) technologies provide a way to expand and enhance our innate cognitive abilities and enable us to accomplish a range of tasks, from simple (e.g., company revenue assessment) to intractable (e.g., air traffic control). However, depending on the timing and context, we may have a greater or lesser ability to make rapid and effective decisions. Yet our ability to do so based on fast-flowing data streams may be decisive. The project \href{https://symbiotik-infovis.eu/}{SYMBIOTIK} proposes a user-centric approach guided by awareness and emotion-sensing capabilities~\cite{leiva2023}, inspired by symbiosis principles where humans and machines cooperate and co-evolve to support decision-making processes. To fulfill this goal, we must define what “awareness” is and how to implement it computationally. Yet researchers have shed little light on the issue of the origins of our subjective experiences, also known as the “hard problems” \cite{chalmers2017hard}. Much progress has been made with what is described as “easy problems,” which relate to access-to-information issues, like understanding our ability to categorize environmental stimuli and the deliberate control of behavior. Solving the hard problems depends on developing a greater understanding of human working memory. In this regard, according to Humphrey \cite{humphrey1984consciousness}, awareness has a strong social dimension that cannot be ignored. Humans have existed in social groups for tens of thousands of years and thus had to predict, understand, and manipulate the behavior of others. Also, according to Baars~\cite{baars1993cognitive}, awareness is the main component of our cognitive processing system, responsible for functions such as adaptation and learning, decision-making, self-monitoring, and self-maintenance. Inspired by this prior body of research, we propose that awareness is a latent property of a properly organized system. In addition, we argue that it is necessary to consider social and collaborative aspects of user-AI interaction with InfoVis systems, orchestrating highly complex and collaborative ways of working while augmenting our visual functions to empower our cognitive functions. Taken together, these innovations will lead to better approaches for shaping the future of adaptive interfaces~\cite{abrahao2021}. In the context of robotics, these innovations can potentially change how people interact with machines in a more natural and user-friendly way. We further need to incorporate basic elements of social functioning (or even priming) by mean of an altruistic, multi-agent environment, as well as emotion recognition capabilities (via cross-modal communication) for establishing the sensing and the affective, which play a fundamental role in our social interactions within humans as well as within machines.

\section{ASTOUND: Improving social competences of virtual agents through artificial consciousness based on the Attention Schema Theory}
The ASTOUND project aims to provide an integrative approach to awareness engineering to establish artificial consciousness in machines. The overarching goal is to develop an AI architecture for Artificial Consciousness based on the Attention Schema Theory (AST), which reconciles some of the current cognitive neuroscience theories of consciousness, stating that subjective awareness results from constructing an internal model of the ``state of attention''. Then, implementing the developed architecture into a contextually aware virtual agent (chatbot) to verify the hypothesis that an artificial consciousness based on AST will unambiguously improve performance in a task of natural language understanding. In this path, we are also targeting to define novel ways to measure the presence and level of consciousness in both humans and machines.

In the context of ASTOUND, we use awareness and artificial consciousness mostly as interchangeable terms, both referring to the quality of a conversational system that allows a coherent and contextualized discussion, self-regulation, and ensuring both short- and long-term interactions with human participants. We relate artificial consciousness to the capability of agents to predict where their own attention and the attention of others is going to focus next. That capability elicits a representation of the activity of attending to something, and that representation enables a form (e.g., verbal) of reporting about the attending activity itself. Here, the proposed architecture will feature an appropriate language registry, adaptation to the interlocutor, and long-term coherence. In addition, it provides larger capabilities for explainability, personalization, multimodal interaction, and understanding.

Currently, conversational systems (chatbots) are a trending area of research, development, and innovation (including robotics as part of the embodiment integration of such systems), as they allow natural interactions, understanding, and reasoning capabilities, as well as provide explanations over actions and environment. The output of the ASTOUND project will increase such capabilities thanks to our research in multimodal processing (text, audio, and image), contextualization, accessibility (especially for deaf people), and automatic evaluation of such systems. We expect a relevant impact, especially in the robotics domain of Cobots, due to the capability of models enriched with an Attention Controller based on an Attention Schema to generate and maintain a model of the attention of others as a proxy to their goals and intentions. In other terms, collaborative robots will be much more effective (and safe) when they are provided with the capability to monitor and correctly predict the objectives of their human partners. In addition, we are working on ethical aspects to make the chatbot compliant with the EU AI act to guarantee a high-quality and safe interaction with users while also being meaningful. Finally, the ASTOUND project targets a high impact as we are building on top of state-of-the-art models and technologies (e.g., Transformer-based pre-trained models) and focused on existing problems such as contextualization, controllability, evaluation, and explanation.

\section{SymAware: Symbolic logic framework for situational awareness in mixed autonomy}
A substantial increase in the deployment of both aerial and ground vehicles with high autonomy is anticipated, necessitating human involvement only in specific circumstances. These vehicles are expected to operate not only in controlled environments but also in dynamic outdoor settings for transportation, logistics, and mobility. In current unpredictable real-world conditions, human situational awareness (SA), risk perception, coordination, and decision-making ensure safety and resilience. However, as operations become more autonomous, relying on the experience of human operators becomes less feasible. In the existing literature, human-based SA models explain how these situations may change over time, indicate who may be affected, and explain how it may be controlled \cite{breakwell2014psychology,endsley1995toward}. Nevertheless, existing robotic agents, including cars, drones, and care assistants, must make decisions based on information beyond what can currently be incorporated into the human-based models of SA.

\href{https://www.symaware.eu/}{SymAware}  addresses the fundamental need for a new conceptual framework for SA in multi-agent systems (MASs) by providing structural and formal modeling of awareness in its various dimensions, sustaining awareness by learning in social contexts, quantifying risks based on limited knowledge, and formulating risk-aware negotiation of task distributions. This framework is aligned with the internal models and specifications of robot agents, facilitating the safe concurrent operation of autonomous agents and humans working together.

\begin{figure}[t]
    \centering
    \includegraphics[width = \textwidth, trim = {0 20 0 0}]{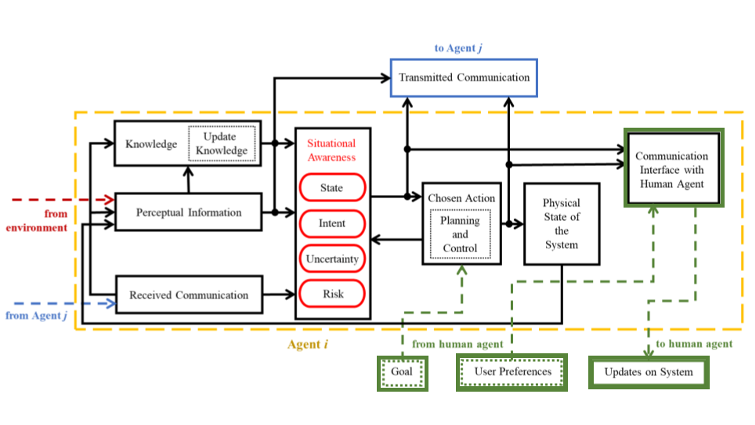}
    \caption{The architecture of an awareness-enabled MAS proposed in \cite{tanevska2023communicating} as part of the project SymAware. The architecture provides insights into the internal structure of an awareness-enabled agent $i$, which interacts with another awareness-enabled agent $j$.}
    \label{fig:SymAware}
\end{figure}

To date, we have proposed a novel architecture for MASs, providing a comprehensive framework that incorporates knowledge, perception, communication, situational awareness, planning, and control to enable effective coordination and functioning of the MAS \cite{tanevska2023communicating}. The first version of the proposed architecture, shown in Fig. \ref{fig:SymAware}, consists of the knowledge component, estimation and perception components, communication modules, situational awareness block, planning and control block, and the physical system. The SA block comprises four elements: (i) state, indicating the current system state combined with environmental information like object presence or obstacles; (ii) intent, deducing other artificial or human agents' intentions by communication or by observing and interpreting their actions to understand their goals; (iii) uncertainty, managing inherent uncertainty and incomplete information in the environment and perception module; and (iv) risk, evaluating potential risks by detecting obstacles, anomalies, or environmental changes to identify threats. 

The implementation and validation of the SymAware approach in awareness engineering will involve two specific scenarios: (i) modeling, simulating, and assessing risks in traffic management for unmanned aircraft systems used in drone operations within urban environments, accounting for potential disturbances and hazards and (ii) simulating and analyzing higher levels of autonomous vehicles, ensuring both functional safety in the face of failures and the Safety of the Intended Functionality.

\section{CAVAA: Counterfactual assessment and valuation for awareness architecture}
The CAVAA project is built on the hypothesis that awareness serves survival in a world governed by hidden states, to deal with the “invisible”, from unexplored environments, projections of the future, to social interaction that depends on the internal states of other agents, conventions, and norms. CAVAA proposes that awareness is the acuity of experience of the conscious scene, reflecting a constructed virtual world, a hybrid of perceptual evidence, memory states, and inferred “unobservables” extended in space and time. Consciousness, in turn, is seen as the coherent experience that results from the large-scale integration of parallel subconscious processing of perception, affect, memory, cognition and action along the neuroaxis supported by a dedicated transient memory system that supports unification, virtualisation, norm extraction and valuation. The CAVAA project builds on the Distributed Adaptive Control theory of mind and brain and will realise this theory of awareness instantiated as an integrated computational architecture and its components to explain awareness in biological systems and engineer it in technological ones. It will realise the underlying computational components of perception, memory, virtualisation, simulation, and integration, embody the architecture in robots and artificial agents, and validate it across a range of use cases involving the interaction between multiple humans and artificial agents, using accepted measures and behavioural correlates of awareness. Use cases will address robot foraging, social robotics, computer game benchmarks and human-generated decision trees in a health coach. These benchmarks will focus on resolving trade-offs, e.g. between search efficiency and robustness, and assess the acceptance of human users of aware technology. CAVAA’s awareness engineering is accompanied by an ethics framework towards human users and aware artefacts in the broader spectrum of trustworthy AI, considering shared ontologies, intention complementarity, behavioural matching, empathy, relevance of outcomes, reciprocity, counterfactuals and projections towards new future scenarios, and to predict the impact of choices. CAVAA will deliver a better user experience because of its explainability, adaptability, and legibility. CAVAA’s integrated framework redefines how we look at the relationship between humans, other species and smart technologies because it makes the invisible visible.

\section{EMERGE: Emergent awareness from minimal collectives}
\begin{figure}[tb]
    \centering
    \includegraphics[width = .85\textwidth]{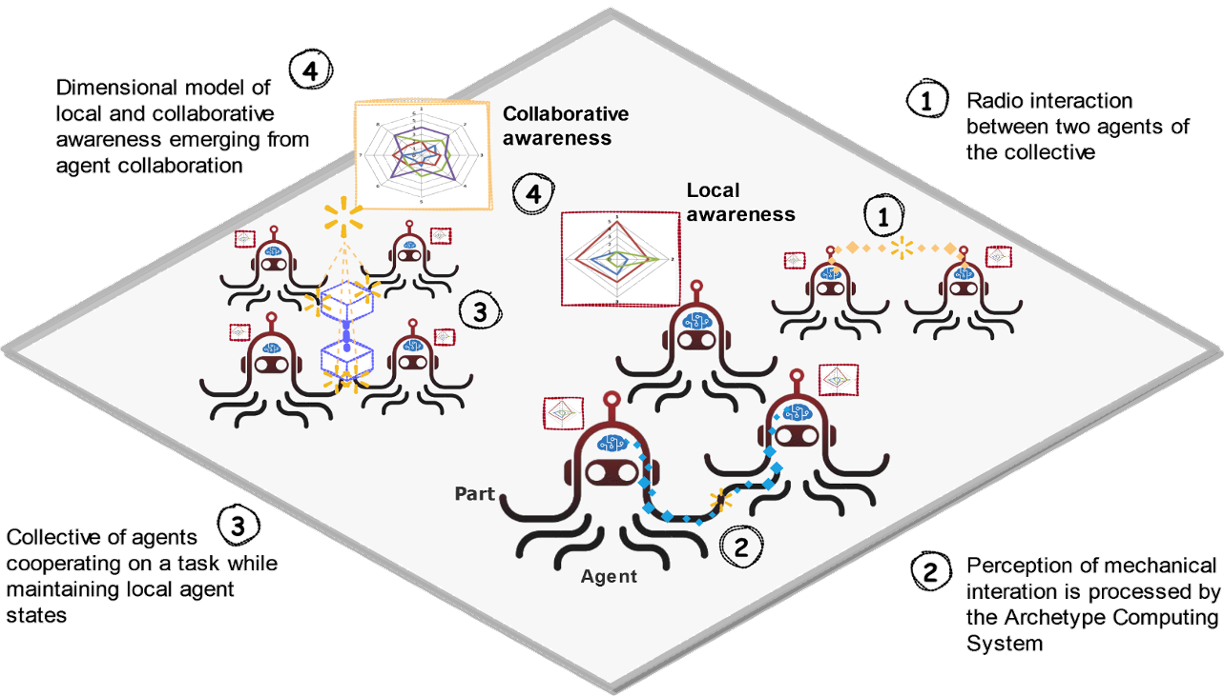}
    \caption{The project EMERGE deals with natural and artificial collaborative awareness, emerging from the interaction of bodies and brains of multiple heterogeneous agents. The figure showcases this idea in the context of swarms of soft robots, which are one of the targeted application domains for the activities of this project.
    \vspace{-0.5cm}
    }
    \label{fig:EMERGE}
\end{figure}

The EMERGE project \cite{bacciu2024emerge} aims to develop a novel concept of collaborative awareness for collectives of minimal artificial agents. In particular, the exploration delves into how collectives of simple agents can develop a representation of their mutual existence, environment, and cooperative behavior toward realizing shared tasks and goals. EMERGE is built on a scenario of artificial beings with no shared language and constrained individual capabilities, which nevertheless leads to high-complexity behaviors at the collective level. EMERGE delivers a philosophical, mathematical, and technological framework that enables us to know how and where to allocate awareness to achieve a goal through the collective optimally. To this end, EMERGE uses nonlinear dynamical system theory to establish, analyze, implement, and test a new AI framework (the Archetype Computing System) that will make collaborative awareness emerge from the interplay of atomic units of local awareness (see Fig. \ref{fig:EMERGE}).

In this project, we identify and characterize a spectrum of global awareness that we call collaborative awareness, which is capable of harmonizing multiple local states belonging to the single agents of the collective in a single coherent global awareness for the ensemble. We intend collaborative awareness as an emergent process supporting complex, distributed, and loosely coupled systems capable of high degrees of collaboration, self-regulation, and interoperability without pre-defined protocols. EMERGE definition of collaborative awareness follows a dimensional model, which allows for the coexistence of multiple awareness aspects (temporal, spatial, bodily, metacognitive, …) existing at different degrees in either fully dependent or more dependent manners.

EMERGE targets the realization of the next generation of self-regulating cyber-physical systems, leveraging collaborative awareness to deliver services over loosely coupled collectives of virtual/physical entities. As such, the project is expected to directly impact IoT applications, smart cities and transportation, microservice-based ICT systems, biomedical nanodevices, and robotics. The latter, in particular, is the primary target application for EMERGE, which explicitly aims to develop three robotic use cases: (i) modular soft robots, as an example of a physically distributed collective where the body needs to self-organize to account for the dynamic addition of components; (ii) robotic swarms, as an example of large scale minimal collective where agents need coordination to achieve a collaborative goal; and (iii) cobots, as a closer-to-market use case where interoperability is currently a significant barrier. We also expect to generate a second stream of impacts by building on the Archetype Computing System to develop neuromorphic and morphological computing systems for green AI and edge AI applications.

\section{METATOOL: Robots inventing tools}
The METATOOL project investigates how robots can develop cognitive abilities to invent tools as ancient humans did around three million years ago. Archaeology, neuroscience, and robotics join forces to shed light on the technological leaps that our ancestors achieved and to develop novel technology inspired by human metacognition and awareness.  Hence, in the same way, tool invention was an outstanding technological milestone in human history, robots inventing tools will be a major milestone in science and technology.

 \begin{figure}[t]
    \centering
    \includegraphics[width = \textwidth]{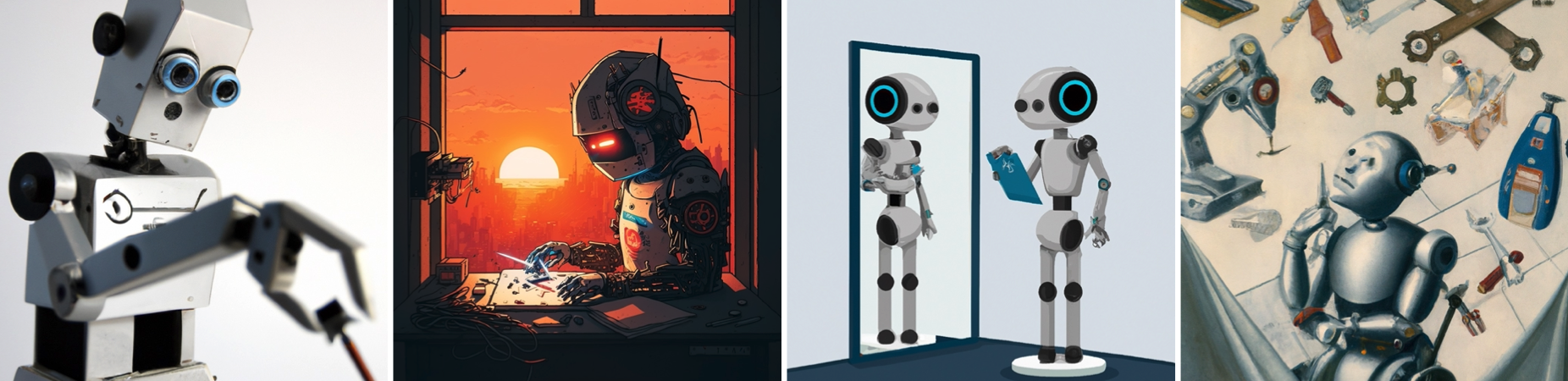}
    \caption{The artificial awareness generated by METATOOL will make robots able to use and invent tools, have self-awareness, and attain operationalizable abstraction.}
    \label{fig:METATOOL}
\end{figure}
Research in METATOOL aims to endow current artificial intelligence systems and robots with a deeper understanding of tools and the capability to monitor and control their actions when interacting with the environment by means of tool-extended bodies (see Fig. \ref{fig:METATOOL}). This implies a mission-oriented understanding of sensory flows from outside and from inside the robot in relation to the performance of a physical task. We identify awareness as the capacity of \emph{real-time understanding of the sensory flows of the agent} \cite{sanz2024awai}, and, inspired by metacognitive theories of cognition, we in particular, focus on the monitoring of the \emph{uncertainty} associated with the cognitive process driving the performance of a task. Thus, analogous to humans, robots would be able to make informed decisions/actions to select, discover, and innovate tools, taking into account both $i)$ the task performance and $ii)$ its self-confidence. We explore the use of a Bayesian computational model originated in neuroscience, \emph{i.e.}, active inference, to deploy this uncertainty monitoring capacity into robots~\cite{lanillos2021active}.
 

By understanding the processes of tool use and invention, we can create robots that are more effective, resilient, and adaptive in open, changing environments. Robots having an intrinsic capability of using objects around them or creating new objects to extend the robot's capability will also relieve engineers from the complex task of modeling all environmental potential affordances. Furthermore, this deeper understanding of the relation between the physical realization of the robot and the task that the robot shall perform will also enable robots to better cooperate with other robots or humans.

\section{SUST(AI)N: Smart building sensitive to daily sentiment}
The SUST(AI)N project aims to provide awareness in smart buildings by means of precision-sensing AI processing shared across devices (i.e., distributed intelligence) and leveraging the flexibility and implementation efficiency of reconfigurable hardware. The integration of these interconnected, self-monitoring systems will enable probabilistic reasoning and, ultimately, the achievement of awareness. Furthermore, zero-energy buildings are sought through energy autonomy in the devices and nodes using energy-efficient sensing, processing, and communication, as well as energy harvesting circuits.
In this project, awareness is defined as the capability of the distributed system, which can be seen as a form of connected superorganism, to reach C2 consciousness through distributed, probabilistic reasoning over possible choices of the learning model, data sources, and node configuration. In particular, we will follow a hierarchical learning model in which the sensor sources, context classes, node configurations, and individual splits in the distributed learning are optimized. Via this hierarchical meta-learning, the system flexibly adapts to the demands and constraints of a building context, achieving self-monitoring and self-healing capabilities and accurate sensing of the group sentiment. 

This project will have a direct impact on tertiary buildings, aiming to make them compliant with the regulatory decree 2010/31/EU (European Parliament), which stipulates the necessity for automation control systems by 2025. The outputs of SUST(AI)N may also have an impact in the field of collaborative/swarm robotics and other instances of multi-agent systems where a distributed form of self-aware intelligence may enhance performance, reliability, and adaptation of the whole system. Particularly, the multi-layered node intelligence, including a node-level intelligence to gather and process local information and a meta-level intelligence layer to decide which information to consider and which machine learning model structure to utilize, could be applied to robots to perform collaborative tasks in an optimal way. Another possible application would be the distributed control of multi-body robots, where each node can be associated with a body part, and the meta-level intelligence can optimally decide how to aggregate the information coming from the various body parts. Energy harvesting can add some degree of energy autonomy to the robots. 

\begin{figure}[t]
    \centering
    \includegraphics[width = 0.85\textwidth, trim = {0 10 0 0}]{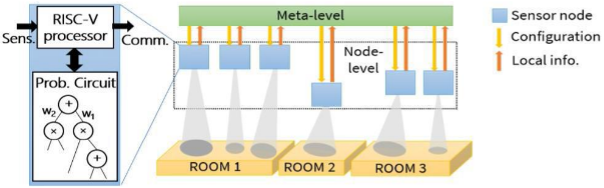}
    \caption{The adaptive awareness that will be developed in SUST(AI)N combines local information across distributed nodes in an aware building scenario.}
    \label{fig:SUSTAIN}
\end{figure}

To date, the project has already developed a RISC-V processor (see Fig. \ref{fig:SUSTAIN}), exploiting probabilistic circuits that will be utilized for the control of the node, an accelerator, and federated learning schemes leveraging various forms of social learning, including both opaque (i.e., neural networks) and transparent models (i.e., based on decision trees) that can also be executed on devices with severely limited computational resources. In addition, indoor light energy harvesting has been explored to power the nodes using several types of solar cells.

\section{VALAWAI: Value-aware AI}
The VALAWAI project is developing tools and methodologies to make AI applications more ethically guided and, hence, trustworthy. More specifically, it aims to infuse AI applications with mechanisms that make them able to abide by human values and be value-aware so that they can recommend or explain behavior in terms of its underlying values. Values are principles of behavior, such as not harming others, speaking the truth, or respecting other opinions. They are reflected in norms that are behavioral patterns designed to satisfy the group's values. Examples of norms are the community rules of social media, the medical protocols adopted by physicians in a hospital for making end-of-life decisions, or the constraints built into a social robot to avoid unacceptable behavior by the robot or its user. General norms for the use of AI are now being translated into legal frameworks, such as the AI Act, although shaping and applying these frameworks to concrete cases is still a work in progress. Obviously, the behavior of robots, particularly social robots, is expected to abide by the values and norms of our society, and so the issue of how to deal with values is critical for the acceptable deployment of robots, particularly in a home context. 

Value-awareness goes beyond value-alignment. Generally speaking, awareness pertains to systems that construct representations of the relevant aspects of a situation and can access these representations in order to deliberatively reason about a course of action and explain why a particular behavior or decision was chosen or why not. For example, situation awareness in the context of firefighting means that firefighters develop a clear understanding of the layout of the building, the inhabitants, the materials on fire, ways to extinguish the fire, etc. Based on these representations, they decide on the best course of action and are able to explain why they followed that course. By analogy, value-aware AI means that the AI system has a representation of the values and norms relevant to a particular situation and uses that to make or recommend decisions and justify them. 

To achieve value-aware AI, the VALAWAI project develops a toolbox with components for representing values and norms and for linking them to concrete situations and decision-making processes. The project valides this toolbox in three use cases: medical protocols, social media, and social robots. The toolbox is generic and can support a variety of functions related to values and norms: alert users when they transgress norms, help a group to communicate and coordinate norms, determine whether certain norms help to achieve certain values and explain behavior in terms of norms. The VALAWAI toolbox, particularly the cognitive architecture contained in it, can also be used for achieving awareness of aspects other than values and norms and has, therefore, a broad utility for robotics in general.

\section{Discussion and Conclusions}

This paper summarized ongoing and diverse activities around the concept of synthetic awareness that have been driven by the EIC Pathfinder Challenge ``Awareness Inside'' call funded by the European Union through the Horizon Europe program. This diversity not only enriches our understanding but also poses significant challenges in defining and achieving a unifying concept of awareness that is applicable across different robotic applications.
For example, the very definition of awareness varies strongly from project to project, with its position towards consciousness ranging from direct opposition (EMERGE) to almost a synonym of hard consciousness (ASTOUND). While in the context of SymAware, awareness is emphasized as the ability of agents in a MAS to recognize and comprehend external stimuli, especially in scenarios involving the presence of human beings, METATOOL awareness focuses on the neuroscience-derived idea of uncertainty monitoring. In both projects, awareness involves perceiving, adapting, coordinating, communicating, and making informed decisions while ensuring safety and resilience in a dynamic environment. 

With this paper, we represented this blooming field of research with the goal of, on the one hand, making a step towards converging on a common framework or language and, on the other involving the whole robotics and AI community in this discussion.

Of the eight projects, five are specifically focused on robotics-related experiments. Namely, METATOOL investigates robots capable of self-evaluation and tool invention; SymAware holds relevance for various applications, encompassing both industrial and domestic robotic systems, with a specific emphasis on autonomous vehicles and air-traffic control systems; CAVAA utilizes social robots like MiRo-e for its tests; EMERGE explores various robotic applications; and VALAWAI showcases its ideas through domestic social robots.
Interestingly, the authors of this paper who are involved in the remaining three projects also believe in applying their innovation to robotics. More specifically, the SYMBIOTIK and ASTOUND innovation could transform social robotics. The SYMBIOTIK project could enhance human-robot interactions by integrating awareness and emotion-sensing capabilities into robotics, making these interactions more natural and user-friendly. Developing artificial consciousness based on the Attention Schema Theory, the ASTOUND project will enhance the performance of natural language understanding and provide collaborative robots with the capability to effectively and safely interact with human partners by predicting their intentions and objectives.
Finally, in line with the goals of EMERGE, the SUST(AI)N project could advance collaborative and swarm robotics, as well as multi-agent systems, by developing a distributed form of self-aware intelligence, enhancing system performance, reliability, and adaptation through a hierarchical learning model and energy-efficient, self-sufficient technologies. 

Ultimately, all the authors of this paper believe that incorporating awareness components in technology could make systems more resilient, adaptable, and human-centric. We believe that understanding awareness will allow AI systems to grasp better and respond to various situations. Synthetic awareness will advance AI towards more coherent, adaptive, and self-evolving behavior.

%
%
\printbibliography

\end{document}